\documentclass[11pt]{article}
\usepackage{emnlp2016}
\usepackage{times}
\usepackage{latexsym}
\usepackage{amssymb}
\usepackage{graphicx}
\usepackage{multirow}
\usepackage{textcomp}
\usepackage{url}
\usepackage{color,comment}

\input{./Definitions.tex}
\emnlpfinalcopy 

\title{Bi-directional Attention with Agreement for Dependency Parsing}
\author{Hao Cheng \quad Hao Fang \\
	University of Washington \\
	\small{\tt{\{chenghao,hfang\}@uw.edu}} \\
\And Xiaodong He \quad  Jianfeng Gao \quad  Li Deng \\
Microsoft Research \\
\small{\tt{\{xiaohe,jfgao,deng\}@microsoft.com}}}

\date{}

\begin{document}

\newcommand{\unk}{\scalebox{0.85}{\ensuremath{<\!\!/{\rm unk}\!\!>}}}

\maketitle

\begin{abstract}
	We develop a novel bi-directional attention model for dependency parsing, which
	learns to agree on headword predictions from the forward and backward parsing directions.
	The parsing procedure for each direction is formulated as sequentially querying 
	the memory component that stores continuous headword embeddings.
	The proposed parser makes use of {\it soft} headword embeddings,
	allowing the model to implicitly capture high-order parsing history without
	dramatically increasing the computational complexity.
	We conduct experiments on English, Chinese, and 12 other languages from the
	CoNLL 2006 shared task, showing that the proposed model achieves
	state-of-the-art unlabeled attachment scores on 6 languages.\footnote{Our software and models are available at \url{https://github.com/hao-cheng/biattdp}.}
\end{abstract}

\section{Introduction}
\label{sec:intro}

Recently, several neural network models have been developed for efficiently
accessing long-term memory and discovering dependencies in sequential data.
The memory network framework has been studied in the context of question
answering and language modeling \cite{Weston2015ICLR,Sukhbaatar2015NIPS},
whereas the neural attention model under the encoder-decoder framework has been
applied to machine translation \cite{Bahdanau2015ICLR} and constituency parsing
\cite{Vinyals2015NIPS}.
Both frameworks learn the latent alignment between the source and target sequences,
and the mechanism of attention over the {\it encoder} can be viewed as a soft
operation on the {\it memory}.
Although already used in the encoder for capturing global context information
\cite{Bahdanau2015ICLR}, the bi-directional recurrent neural network (RNN) has
yet to be employed in the decoder.
Bi-directional decoding is expected to be advantageous over the previously
developed uni-directional counterpart, because the former
exploits richer contextual information.
Intuitively, we can use two separate uni-directional RNNs where each one constructs
its respective attended encoder context vectors for computing RNN hidden states.
However, the drawback of this approach is that the decoder would often produce
different alignments resulting in discrepancies for the forward and backward
directions.
In this paper, we design a training objective function to enforce
attention agreement between both directions, inspired by the
alignment-by-agreement idea from \newcite{Liang2006NAACL}.
Specifically, we develop a dependency parser (BiAtt-DP) using a bi-directional
attention model based on the memory network.
Given that the golden alignment is observed for dependency parsing in the
training stage, we further derive a simple and interpretable approximation for
the agreement objective, which makes a natural connection between the latent and
observed alignment cases.
\vspace{-0.2ex}

The proposed BiAtt-DP parses a sentence in a linear order via sequentially
querying the memory component that stores continuous embeddings for all headwords.
In other words, we consider all possible arcs during the parsing.
This formulation is adopted by graph-based parsers such as the MSTParser
\cite{McDonald2005EMNLP}.
The consideration of all possible arcs makes the proposed BiAtt-DP different from many recently
developed neural dependency parsers
\cite{Chen2014EMNLP,Weiss2015ACL,Alberti2015EMNLP,Dyer2015ACL,Ballesteros2015EMNLP},
which use a transition-based algorithm by modeling the parsing procedure as a
sequence of actions on buffers.
Moreover, unlike most graph-based parsers 
which may suffer from high computational complexity when utilizing high-order
parsing history \cite{McDonald2006EACL}, the proposed BiAtt-DP can implicitly
inject such information into the model while keeping the computational complexity
in the order of $\mathcal{O}(n^2)$ for a sentence with $n$ words.
This is achieved by feeding the RNN in the query component with a {\it soft}
headword embedding, which is computed as the probability-weighted sum of all
headword embeddings in the memory component.


To the best of our knowledge, this is the first attempt to apply memory network
models to graph-based dependency parsing. 
Moreover, it is the first extension of neural attention models from
uni-direction to multi-direction by enforcing agreement on alignments. 
Experiments on English, Chinese, and 12 languages from the CoNLL 2006
shared task show the BiAtt-DP can achieve competitive parsing
accuracy with several state-of-the-art parsers.
Furthermore, our model achieves the highest unlabeled attachment score (UAS) on
Chinese, Czech, Dutch, German, Spanish and Turkish. 

\section{A MemNet-based Dependency Parser}
\label{sec:memnet_dp}
The proposed parser first encodes each word in a sentence by continuous
embeddings using a bi-directional RNN, and then performs two types of
operations, 
\ie 1) headword predictions based on bi-directional parsing history and 2) the
relation prediction conditioned on the current modifier and its
\textit{predicted} headword both in the embedding space.
In the following, we first present how the token embeddings are constructed.
Then, the key components of the proposed parser, \ie the memory component and
the query component, are discussed in detail.
Lastly, we describe the parsing algorithm using a bi-directional attention model
with agreement.

\subsection{Token Embeddings}
In the proposed BiAtt-DP, the memory and query components share the same token
embeddings.
We use the notion of additive token embedding as in \cite{Botha2014ICML} to
utilize the available information about the token, e.g., its word form, lemma,
part-of-speech (POS) tag, and morphological features. 
Specifically, the token embedding is computed as
\vspace{-2.5ex}

{\small
\begin{align*}
	{\bf E}^{\rm form} \evec^{\rm form}_i 
	+ {\bf E}^{\rm pos} \evec^{\rm pos}_i 
	+ {\bf E}^{\rm lemma} \evec^{\rm lemma}_i 
	+ \cdots,
\end{align*}}%
where $\evec_i$'s are one-hot encoding vectors for the $i$-th word, and ${\bf E}$'s are parameters
to be learned that store the continuous embeddings for corresponding feature.
Note those one-hot encoding vectors have different dimensions, depending on individual
vocabulary sizes, and all ${\bf E}$'s have the same first dimension but different 
second dimension.
The additive token embeddings allow us to easily integrate a variety of
information. 
Moreover, we only need to make a single decision on the dimensionality of the
token embedding, rather than a combination of decisions on word embeddings and
POS tag embeddings as in concatenated token embeddings used by
\newcite{Chen2014EMNLP}, \newcite{Dyer2015ACL} and \newcite{Weiss2015ACL}.
It reduces the number of model parameters to be tuned, especially when lots of
different features are used.
In our experiments, the word form and fine-grained POS tag are always used,
whereas other features are used depending on their availability in the dataset.
All singleton words, lemmas, and POS tags are replaced by special tokens.

The additive token embeddings are transformed into another space before they are
used by the memory and query components, \ie
\begin{align*}
	\xvec_i = {\rm LReL} \left[
		{\bf P} \left(
			{\bf E}^{\rm form} \evec^{\rm form}_i + \cdots 
		\right) 
	\right],
\end{align*}
where ${\bf P}$ is the projection matrix and is shared by the memory and query
components as well.
The activation function of this projection layer is the leaky rectified linear
(LReL) function \cite{Mass2013ICML} with 0.1 as the slope of the negative part.
In the remaining part of the paper, we refer to $\xvec_i\in\RR^p$ as the token
embedding for word at position $i$. 
Note the subscript $i$ is substituted by $j$ and $t$ for the memory and query
components, respectively.

\subsection{Components}
\label{ssec:model_comp}
\begin{figure}[!t]
	\centering
	\includegraphics[width=0.45\textwidth]{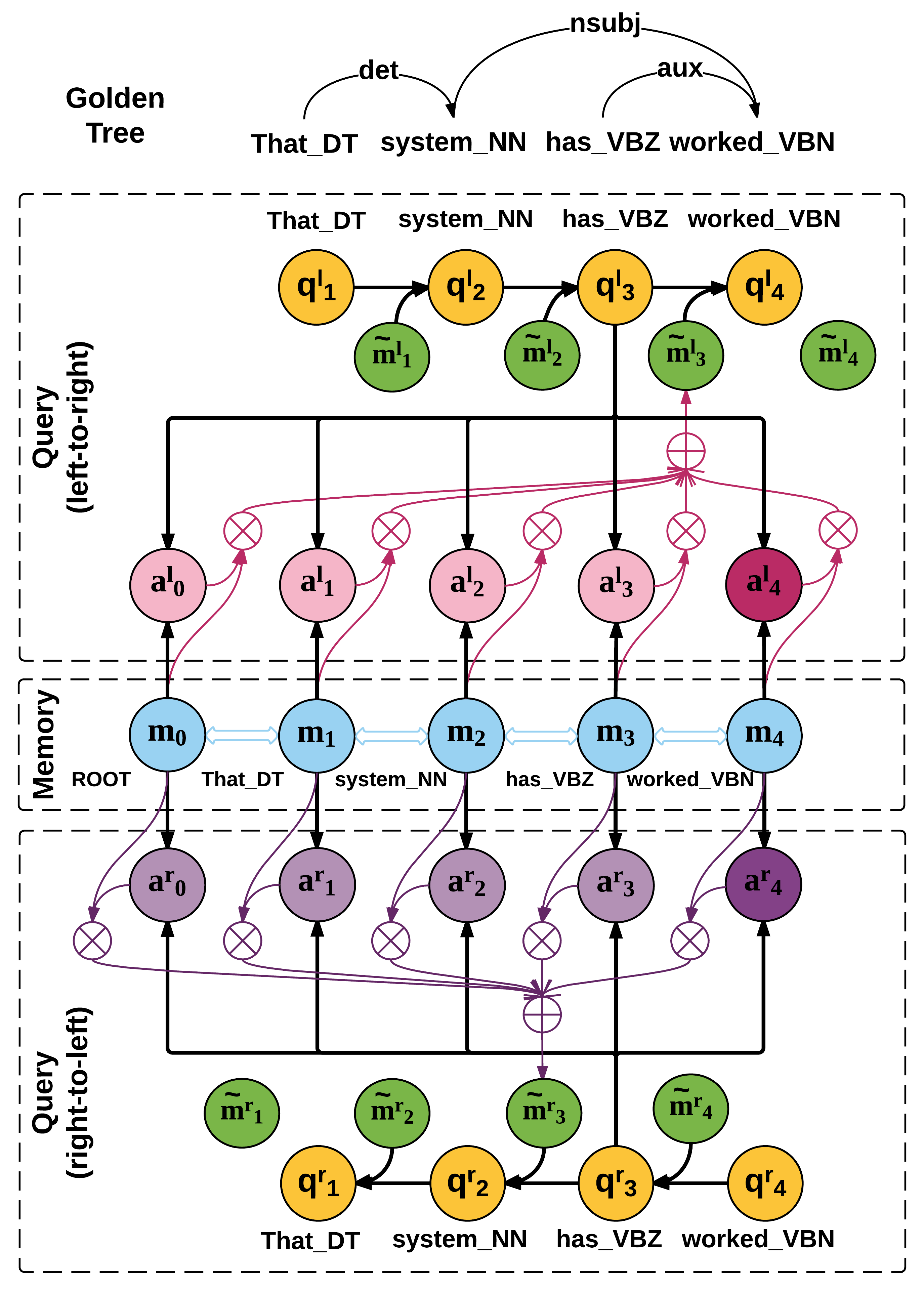}
	\caption{The structure of the BiAtt-DP.
		The figure only illustrates the parsing process at the time step for \texttt{has}.
		Blue and yellow circles are memory and query vectors, respectively.
		Red and purple circles represent headword probabilities predicted from
		corresponding query components.
		Green circles represent soft headword embeddings.
		Black arrowed lines are connections carrying weight matrices.
		$\otimes$ and $\oplus$ indicate element-wise multiplication and addition,
		respectively.
		For simplicity, we ignore the token embedding $\xvec_t$ connected to the
	RNN hidden layers $\mvec_j$, $\qvec^l_t$ and $\qvec^r_t$.}
	\label{fig:memnet_parser}
\end{figure}

As shown in Figure\,\ref{fig:memnet_parser}, 
the proposed BiAtt-DP has
three components, \ie a memory component, a left-to-right query component, and a
right-to-left query component.
Given a sentence of length $n$, the parser first uses a bi-directional
RNN to construct $n + 1$ headword embeddings, $\mvec_0, \mvec_1, \ldots,
\mvec_n\in\RR^e$, with $\mvec_0$ reserved for the \texttt{ROOT} symbol.
Each query component is an uni-directional attention model.
In a query component, a sequence of $n$ modifier embeddings 
$\qvec_1, \ldots, \qvec_n \in \RR^d$ are
constructed recursively by conditioning on all headword embeddings.
To address the vanishing gradient issue in RNNs, we use the gated recurrent
unit (GRU) proposed by \newcite{Cho2014EMNLP}, where an update gate and a reset
gate are employed to control the information flow.
We replace the hyperbolic tangent function in GRU with the LReL function, which
is faster to compute and achieves better parsing accuracy in our preliminary
studies.
In the following, we refer to headword and modifier embeddings as memory and
query vectors, respectively.

\paragraph{Memory Component:}
The proposed BiAtt-DP uses a bi-directional RNN to obtain the memory vectors.
At time step $j$, the current hidden state vector $\hvec^l_j \in
\mathbb{R}^{e/2}$ (or $\hvec^r_j \in \mathbb{R}^{e/2}$) is
computed as a non-linear transformation based on the current input vector
$\xvec_j$ and the previous hidden state vector $\hvec^l_{j-1}$ (or $\hvec^r_{j+1}$),
\ie $\hvec^l_j = {\rm GRU} (\hvec^l_{j-1}, \xvec_j)$
(or $\hvec^r_j = {\rm GRU} (\hvec^r_{j+1}, \xvec_j)$).
Ideally, the recursive nature of the RNN allows it to capture all context
information from one-side, and a bi-directional RNN can thus capture context
information from both sides.
We concatenate the hidden layers of the left-to-right RNN and the right-to-left
RNN for the word at position $j$ as the memory vector 
$\mvec_j = \left[ \hvec^l_j; \hvec^r_j \right]$.
These memory vectors are expected to encode the words and their context
information in the headword space.

\paragraph{Query Component:}
For each query component, we use a single-directional RNN with GRU to obtain the
query vectors $\qvec_j$'s, which are the hidden state vectors of the RNN.
Each $\qvec_t$ is used to query the memory component, returning association 
scores $s_{t, j}$'s between the word at position $t$ and the headword at
position $j$ for $j \in \{0, \cdots, n\}$, \ie
\begin{align}
	s_{t, j} = \vvec^T 
	\phi \left( {\bf C}\mvec_j + {\bf D}\qvec_t \right),
	\label{eq:attention_score}
\end{align}
where $\phi(\cdot)$ is the element-wise hyperbolic tangent function, and ${\bf
C}\in\RR^{h\times e}$, ${\bf D}\in\RR^{h\times d}$
and $\vvec\in\RR^h$ are model parameters.
Then, we can obtain probabilities (aka attention weights), $a_{t, 0}, \cdots,
a_{t, n}$, over all headwords in the sentence by normalizing
$s_{t, j}$'s, using a softmax function
\begin{eqnarray}
	\avec_t = {\rm softmax} (\svec_t).
	\label{eq:prob_headword}
\end{eqnarray}
The {\it soft} headword embedding is then defined as 
$\tilde{\mvec}_t=\sum_{j=1}^n a_{t,j}\mvec_j$.
At each time step $t$, the RNN takes the {\it soft} headword embedding
$\tilde{\mvec}^{l}_{t-1}$ or $\tilde{\mvec}^{r}_{t+1}$ as the input, in addition
to the token embedding $\xvec_t$.
Formally, for the forward case, the $\qvec_t$ can be computed as
$\qvec_t = {\rm GRU} 
\left( \qvec_{t-1}, \left[ \tilde{\mvec_{t}}; \xvec_t \right] \right)$.
Although the RNN is able to capture long-span context
information to some extent, the local context may very easily dominate the
hidden state.
Therefore, this additional soft headword embedding allows the model to access
long-span context information in a different channel.
On the other hand, by recursively feeding both the query vector and the soft
headword embedding into the RNN, the model {\it implicitly} captures high-order
parsing history information, which can potentially improve the parsing accuracy
\cite{Yamada2003IWPT,McDonald2006EACL}. 
However, for a graph-based dependency parser, utilizing parsing history features
is computationally expensive.
For example, an $k$-th order MSTParser \cite{McDonald2006EACL} has
$\mathcal{O}(n^{k+1})$ complexity for a sentence of $n$ words. 
In contrast, the BiAtt-DP {\it implicitly} captures high-order parsing history 
while keeping the complexity in the order of $\mathcal{O}(n^2)$,
\ie for each direction. we compute $n(n + 1)$ pair-wise probabilities $a_{t,j}$
for $t=1, \cdots, n$ and $j = 0, \cdots, n$.

In this paper, we choose to use soft headword embeddings rather than making
hard decisions on headwords.
In the latter case, beam search may potentially improve the parsing accuracy at
the cost of higher computational complexity, \ie $\mathcal{O}(Bn^2)$ with a beam
width of $B$.
When using soft headword embeddings, there is no need to perform beam search.
Moreover, it is straightforward to incorporate parsing history from both
directions by using two query components at the cost of $\mathcal{O}(2n^2)$, which
cannot be easily achieved when using beam search.
The parsing decision can be made directly based on attention weights from the
two query components or further rescored by the maximum spanning tree (MST) search algorithm.

\subsection{Parsing by Attention with Agreement}
\label{ssec:parse_alg}
For the bi-directional attention model, the underlying probability distributions
$\avec^l_t$ and $\avec^r_t$ may not agree with each other.
In order to encourage the agreement, we use the mathematically convenient
metric, \ie the squared Hellinger distance 
$H^2 \left( \avec_t^l || \avec_t^r \right)$, 
for quantifying the distance between these two distributions.
For dependency parsing, when the golden alignment is known during training,
we can derive an upper bound on the latent agreement objective as 
\begin{align*}
	H^2 (\avec_t^l, \avec_t^r)
	\leq
	2 \sqrt{ D(\gvec_t || \avec_t^l)
	+ D(\gvec_t || \avec_t^r)},
\end{align*}
where $D(\cdot||\cdot)$ is the KL-divergence.
The complete derivation is provided in the Appendix~\ref{sec:appendix_ub}.
During optimization, we can safely drop the constant scaler and the square root
operation in the upper bound, leading to the following loss function
\begin{align}
	D(\gvec_t || \avec_t^l) + D(\gvec_t || \avec_t^r)
	= 2 D (\gvec_t || \avec_t^l \odot \avec_t^r),
	\label{eqn:align_obj2}
\end{align}
where $\odot$ indicates element-wise multiplication.
The resulting loss function is equivalent to the cross-entropy loss, which is
widely adopted for training neural networks.

As we can see, the loss function (\ref{eqn:align_obj2}) tries to minimize the
distance between the golden alignment and the intersection of the two
directional attention alignments at every time step.
Therefore, during inference, the headword prediction for the word at time step
$t$ can be obtained as 
\begin{align*}
	\argmax_j \quad \log a^{l}_{t, j} + \log a^{r}_{t, j},
\end{align*}
seeking for agreement between both query components. 
This parsing procedure is also similar to the exhaustive left-to-right
modifier-first search algorithm described in \cite{Covington2001ACM}, but it is
enhanced by an additional right-to-left search with the agreement enforcement.
Alternatively, we can treat $(\log a^{l}_{t, j} + \log a^{r}_{t, j})$ as a score
of the corresponding arc and then search for the MST to
form a dependency parse tree, as proposed in \cite{McDonald2005EMNLP}.
The MST search is achieved via the Chu-Liu-Edmonds algorithm
\cite{Chu1965,Edmonds1967}, which can be implemented in $\mathcal{O}(n^2)$ for
dense graphs according to \newcite{Tarjan1977}.
In practice, the MST search slows down the parsing speed by 6--10\%.
However, it forces the parser to produce a valid tree, and we observe a slight
improvement on parsing accuracy in most cases.

After obtaining each modifier and its \textit{soft} header embeddings, we use a
single-layer perceptron to predict the head-modifier relation, \ie
\vspace{-3ex}

{\small
\begin{align}
	\yvec_t = {\rm softmax} \left( 
	{\bf U} \left[ \tilde{\mvec}^{l}_t; \: \tilde{\mvec}^{r}_t \right]
	+ {\bf W} \left[ \qvec^{l}_t; \: \qvec^{r}_t \right] 
	\right),
	\label{eq:prob_relation}
\end{align}}%
where $y_{t, 1}, \cdots, y_{t, m}$ are the probabilities of $m$ possible
relations, and ${\bf U}\in\RR^{m\times 2e}$ and ${\bf W}\in\RR^{m\times 2d}$ are
model parameters.

\section{Model Learning}
For the $t$-th word (modifier) $w_t$ in a sentence of length $n$, 
let $H^{l}_t$ and $H^{r}_t$ denote random variables representing the predicted
headword from forward (left-to-right) and backward (right-to-left) parsing
directions, respectively.
Also let $R_t$ denote the random variable representing the dependency
relation for $w_t$.
The joint probability of headword and relation predictions can be written as
\begin{align}
	& \quad P ( R_{1:n}, H_{1:n}^l, H_{1:n}^r | w_{1:n} ) \nonumber \\
	&= \prod_{t=1}^n P ( R_t | w_{1:n} ) 
	P ( H_t^l | w_{1:n} ) 
	P ( H_t^r | w_{1:n} ) \nonumber \\
	&= \prod_{t=1}^n y^l_{t, R_t} \cdot a^l_{t, H^l_t} \cdot a^r_{t, H^r_t}
	\label{eqn:parse_obj}
\end{align}
where at each time step we assume head-modifier relations and headwords from
both directions are independent with each other when conditioned on the global
knowledge of the whole sentence.
Note that the long-span context and high-order parsing history information are
injected when we model $P ( H_t^l | w_{1:n} ), P ( H_t^r | w_{1:n} )$ and $ P (
R_t | w_{1:n} )$, as discussed in Section~\ref{ssec:model_comp}.

As discussed in Section~\ref{ssec:parse_alg}, the model can be trained by
encouraging attention agreement between two query components.
From (\ref{eqn:parse_obj}), we observe that it is equivalent to maximizing the
log-likelihood of the golden dependency tree (or minimizing the
cross-entropy) for each training sentence, \ie 
\begin{align*}
	\sum_{t=1}^n\left(
	\log y_{t, {\rm relation}_t}
	+ \log a^{l}_{t, {\rm head}_t} 
	+ \log a^{r}_{t, {\rm head}_t} 
	\right),
\end{align*}
where $a_{t, j}$ and $y_{t, r}$ are defined in (\ref{eq:prob_headword}) and
(\ref{eq:prob_relation}), respectively, and \texttt{relation}$_t$ and
\texttt{head}$_t$ are golden relation and headword labels, respectively.
%
The gradients are computed via the back-propagation algorithm
\cite{Rumelhart1986Nature}.
Errors of $\yvec_t$ come from the arc labels, whereas there are two source of
errors for $\avec_t$, one 
from the headword labels and the other back-propagated from errors of $\yvec_t$.
We use stochastic gradient descent with the Adam algorithm proposed in
\cite{Kingma2015ICLR}.
The learning rate is halved at each iteration once the log-likelihood of the
dev set decreases. 
The whole training procedure terminates when the log-likelihood decreases for
the second time.
All learning parameters except bias terms are initialized randomly according to
the Gaussian distribution $\mathcal{N}(0, 10^{-2})$.
In our experiments, we tune the initial learning rate with a step size of
0.0002, and choose the best one based on the log-likelihood on the dev set at
the first epoch.
Empirically, the selected initial learning rates fall in the range of $[0.0004,
0.0010]$ for hidden layer size $[128, 320]$, and tend to be larger when using a
smaller hidden layer size, \ie $[0.0016, 0.0034]$ for hidden layer size around
80.
The training data are randomly shuffled at every epoch.

\section{Experiments}
\label{sec:exp_start}

In this section, we present the parsing accuracy of the proposed BiAtt-DP on 14
languages.
We report both UAS and labeled attachment score (LAS), obtained by the
CoNLL-X eval.pl
script\footnote{\url{http://ilk.uvt.nl/conll/software.html}} which ignores
punctuation symbols.
The headword predictions are made through the MST search, which slightly
improves both UAS and LAS (less than $0.3\%$ absolutely).
Overall, the proposed BiAtt-DP achieves competitive parsing accuracy on all
languages as state-of-the-art parsers, and obtains better UAS in 6 languages.
We also show the impact of using POS tags and pre-trained word embeddings.
Moreover, different variants of the full model are compared in this section.

\subsection{Data}
\label{sec:data}

We work on the English Treebank-3 (PTB) dataset \cite{LDC99T42},
the Chinese Treebank-5.1 (CTB) dataset \cite{LDC2005T01}, and 12 other languages
from the CoNLL 2006 shared task \cite{CONLL2006}.
For PTB and CTB datasets, we use exactly the same setup as in
\cite{Chen2014EMNLP,Dyer2015ACL}.
Specifically, we convert the English and Chinese data using the Stanford parser
v3.3.0 \cite{Marneffe2006LREC} and the Penn2Malt tool \cite{Zhang2008EMNLP},
respectively.

For English, POS tags are obtained using the Stanford POS tagger v3.3.0
\cite{Toutanova2003NAACL}, whereas for Chinese, we use gold segmentation and POS
tags.
When constructing the token embeddings for English and Chinese, both the
word form and the POS tag are used.
We also initialize ${\bf E}^{\rm form}$ by 
pre-trained word embeddings\footnote{For English, we use the dependency-based
	word embeddings at \url{https://goo.gl/tWke3I} \cite{Levy2014ACL}.
	For Chinese, we pre-train 192-dimension skip-gram embeddings
	\cite{Mikolov2013ICLR} on Chinese Gigawords \cite{LDC2005T14}.}.

For the 12 other languages, we randomly hold out $5\%$ of the training data
as the dev set. 
In addition to the word form and find-grained POS tags, we use extra features
such as lemmas, coarse-grained POS tags, and morphemes when they are available in
the dataset.
No pre-trained word embeddings are used for these 12 languages.

\subsection{Model Configurations}
\label{ssec:model_config}
The hidden layer size is kept the same across all RNNs in the proposed
BiAtt-DP.  
We also require the dimension of the token embeddings to be the same as the
hidden layer size.
Note that we concatenate the hidden layers of two RNNs for constructing
$\mvec_j$, and thus we have $e = 2d$.
The weight matrices ${\bf C}$ and ${\bf D}$ respectively project vectors
$\mvec_j$ and $\qvec_t$ to the same dimension $h$, which is equivalent to $d$. 
For English and Chinese, since the dimension of pre-trained word embeddings are
300, we use $300 \times h$ as the dimension of embedding parameters ${\bf E}$'s.
For the 12 other languages, we use square matrices for the embedding parameters
${\bf E}$'s. 
For all languages, We tune the hidden layer size and choose one according to 
UAS on the dev set.
The selected hidden layer sizes for these languages are:
368 (English),
114 (Chinese), 
128 (Arabic), 
160 (Bulgarian), 
224 (Czech), 
176 (Danish),
220 (Dutch), 
200 (German), 
128 (Japanese), 
168 (Portuguese),
128 (Slovene),
144 (Spanish), 
176 (Swedish), 
and 128 (Turkish).

\begin{table}[t]
	\centering
	{\small
	\begin{tabular}{l|l|cc}
		\hline\hline
		{\bf Type} & \quad\qquad {\bf Method} 
		& {\bf UAS} & {\bf LAS} \\
		\hline
		\multirow{5}{*}{Trans.}
		& C\&M (2014) & 91.8\underline{\:\:} & 89.6\underline{\:\:} \\
		& \protect\newcite{Dyer2015ACL} & 93.2\underline{\:\:} & 90.9\underline{\:\:} \\
		& B\&N (2012)$^\dag$ & 93.33 & 91.22 \\
		& \protect\newcite{Alberti2015EMNLP}$^\dag$ & 94.23 & 92.41 \\
		& \protect\newcite{Weiss2015ACL}$^\dag$ & 94.26 & 92.41 \\
		& \protect\newcite{Andor2016ACL}$^*$ 		& {\bf 94.41} & {\bf 92.55} \\
		\hline
		\multirow{4}{*}{Graph}
		& \protect\newcite{Bohnet2010COLING}$^\dag$ & 92.88 & 90.71 \\
		& \protect\newcite{Martins2013ACL}$^\dag$ & 92.89 & 90.55 \\
		& Z\&M (2014)$^\dag$ & 93.22 & 91.02 \\
		& BiAtt-DP & {\bf 94.10} & {\bf 91.49} \\
		\hline\hline
	\end{tabular}}%
	\caption{Parsing accuracy on PTB test set. 
		Our parser uses the same POS tagger as C\&M (2014) and
		\protect\newcite{Dyer2015ACL}, whereas other parsers use a different POS
		tagger.
		Results with $\dag$ and $*$ are provided in \protect\cite{Alberti2015EMNLP}
		and \protect\cite{Andor2016ACL}, respectively.}
	\label{tab:exp_ptb}
\end{table}

\begin{table}[t]
	\centering
	{\small
	\begin{tabular}{l|cc|cc}
		\hline\hline
		& \multicolumn{2}{c|}{\bf Dev}
		& \multicolumn{2}{c}{\bf Test} \\
		& UAS & LAS & UAS & LAS \\
		\hline
		C\&M (2014) & 84.0 & 82.4 & 83.9 & 82.4 \\
		\protect\newcite{Dyer2015ACL} & 87.2 & {\bf 85.9} & 87.2 & {\bf 85.7} \\
		BiAtt-DP & {\bf 87.7} & 85.3 & {\bf 88.1} & {\bf 85.7} \\
		\hline\hline
	\end{tabular}}%
	\caption{Parsing accuracy on CTB dev and test sets.}
	\label{tab:exp_ctb}
\end{table}

\begin{table*}[t]
	\centering
	{\small
	\begin{tabular}{l|ccccc|ccc}
		\hline\hline
		{\bf Language}
		& \multicolumn{2}{c}{\bf BiAtt-DP}
		& {\bf RBGParser}
		& \multicolumn{2}{c|}{\bf Best Published}
		& {\bf Crossed} & {\bf Uncrossed} & {\bf \%Crossed} \\
		\hline
		Arabic 			& 80.34 			& [68.58]  & 79.95 & {\bf 81.12} 	& (Ma11) & 17.24 & 80.71 & 0.58 \\
		Bulgarian 	& 93.96 			& [89.55]  & 93.50 & {\bf 94.02} 	& (Zh14) & 79.59 & 94.10 & 0.98 \\
		Czech 			& {\bf 91.16} & [85.14]  & 90.50 & 90.32 				& (Ma13) & 81.62 & 91.63 & 4.68 \\
		Danish 			& 91.56 			& [85.53]  & 91.39 & {\bf 92.00} 	& (Zh13) & 73.33 & 91.89 & 1.80 \\
		Dutch 			& {\bf 87.15} & [82.41]  & 86.41 & 86.19 				& (Ma13) & 82.82 & 87.66 & 10.48 \\
		German 			& {\bf 92.71} & [89.80]  & 91.97 & 92.41 				& (Ma13) & 85.93 & 92.90 & 2.70 \\
		Japanese 		& 93.44 			& [90.67]  & 93.71 & {\bf 93.72} 	& (Ma11) & 48.67 & 94.48 & 2.26 \\
		Portuguese 	& 92.77 			& [88.44]  & 91.92 & {\bf 93.03}	& (Ko10) & 73.02 & 93.28 & 2.52 \\
		Slovene 		& 86.01 			& [75.90]  & 86.24 & {\bf 86.95} 	& (Ma11) & 60.11 & 86.99 & 3.66 \\
		Spanish 		& {\bf 88.74} & [84.03]  & 88.00 & 87.98 				& (Zh14) & 50.00 & 88.77 & 0.08 \\
		Swedish 		& 90.50 			& [84.05]  & 91.00 & {\bf 91.85} 	& (Zh14) & 45.16 & 90.78 & 0.62 \\
		Turkish 		& {\bf 78.43} & [66.16]  & 76.84 & 77.55 				& (Ko10) & 38.85 & 79.71 & 3.13 \\
		\hline\hline
	\end{tabular}}%
	\caption{UAS on 12 languages in the CoNLL 2006 shared task
		\protect\cite{CONLL2006}.
		We also report corresponding LAS in squared brackets. 
		The results of the 3rd-order RBGParser are reported in \protect\cite{Lei2014ACL}.
		Best published results on the same dataset in terms of UAS
		among \protect\cite{Pitler2015NAACL}, \protect\cite{ZhangMcDonald2014ACL},
		\protect\cite{Zhang2013EMNLP}, \protect\cite{ZhangMcDonald2012EMNLP},
		\protect\cite{Rush2012NAACL}, \protect\cite{Martins2013ACL},
		\protect\cite{Martins2010EMNLP}, and \protect\cite{Koo2010EMNLP}.
		To study the effectiveness of the parser in dealing with
		non-projectivity, we follow \protect\cite{Pitler2015NAACL}, to compute
		the recall of crossed and uncrossed arcs in the gold tree, as well as the
	percentage of crossed arcs.}
	\label{tab:exp_conll2006}
\end{table*}

\subsection{Results}
\label{sec:result}

We first compare our parser with state-of-the-art neural transition-based
dependency parsers on PTB and CTB.
For English, we also compare with state-of-the-art graph-based dependency
parsers.
The results are shown in Table~\ref{tab:exp_ptb} and Table~\ref{tab:exp_ctb},
respectively.
It can be seen that the BiAtt-DP outperforms all other graph-based parsers on
PTB.
Compared with the transition-based parsers, it achieves better accuracy than
\newcite{Chen2014EMNLP}, which uses a feed-forward neural network, and
\newcite{Dyer2015ACL}, which uses three stack LSTM networks.
Compared with the integrated parsing and tagging models, the BiAtt-DP
outperforms \newcite{Bohnet2012EMNLP} but has a small gap to \newcite{Alberti2015EMNLP}.
On CTB, it achieves best UAS and similar LAS.
This may be caused by that the relation vocabulary size is relatively smaller
than the average sentence length, which biases the joint objective to be more
sensitive to UAS.
The parsing speed is around 50--60 sents/sec measured on a desktop with Intel
Core i7 CPU @ 3.33GHz using single thread.

Next, in Table\,\ref{tab:exp_conll2006} we show the parsing accuracy of the
proposed BiAtt-DP on 12 languages in the CoNLL 2006 shared task, including
comparison with state-of-the-art parsers.
Specifically, we show UAS of the 3rd-order RBGParser as reported in
\cite{Lei2014ACL} since it also uses low-dimensional continuous embeddings.
However, there are several major differences between the RBGParser and the
BiAtt-DP.
First, in \cite{Lei2014ACL}, the low-dimensional continuous embeddings are
derived from low-rank tensors.
Second, the RBGParser uses combined scoring of arcs by including traditional
features from the MSTParser \cite{McDonald2006EACL} / TurboParser
\cite{Martins2013ACL}.
Third, the RBGParser employs a third-order parsing algorithm based on
\cite{ZhangLei2014ACL}, although it also implements a first-order parsing
algorithm, which achieves lower UAS in general.
In Table\,\ref{tab:exp_conll2006}, we show that the proposed BiAtt-DP
outperforms the RBGParser in most languages except Japanese, Slovene, and
Swedish.

It can be observed from Table\,\ref{tab:exp_conll2006} that the BiAtt-DP has
highly competitive parsing accuracy as state-of-the-art parsers.
Moreover, it achieves best UAS for 5 out of 12 languages.
For the remaining seven languages, the UAS gaps between the BiAtt-DP and 
state-of-the-art parsers are within 1.0\%, except Swedish.
An arguably fair comparison for the BiAtt-DP is the MSTParser
\cite{McDonald2006EACL}, since the BiAtt-DP replaces the scoring function for
arcs but uses exactly the same search algorithm.
Due to the space limit, we refer readers to \cite{Lei2014ACL} for results of the
MSTParsers (also shown in Appendix~\ref{sec:appendix_uas_mst}).
The BiAtt-DP consistently outperforms both parser by up to 5\% absolute UAS
score.

Finally, following \cite{Pitler2015NAACL}, we also analyze the performance of
the BiAtt-DP on both crossed and uncrossed arcs.
Since the BiAtt-DP uses a graph-based non-projective parsing algorithm, it is
interesting to evaluate the performance on crossed arcs, which result in the
non-projectivity of the dependency tree.
The last three columns of Table\,\ref{tab:exp_conll2006} show the recall of
crossed arcs, that of uncrossed arcs, and the percentage of crossed arcs in the
test set.  \newcite{Pitler2015NAACL} reported numbers on the same data for
Dutch, German, Portuguese, and Slovene as in this paper.
For these four languages, the BiAtt-DP achieves better UAS than that reported
in \cite{Pitler2015NAACL}.
More importantly, we observe that the improvement on recall of crossed
arcs (around 10--18\% absolutely) is much more significant than that of
uncrossed arcs (around 1--3\% absolutely), which indicates the effectiveness of
the BiAtt-DP in parsing languages with non-projective trees.

\subsection{Ablative Study}
\label{sec:ab_study}

\begin{table}[t]
	\centering
	{\small
	\begin{tabular}{ccccc|cc}
		\hline\hline
		No. & INIT & POS & L2R & R2L & UAS & LAS \\
		\hline
		1 &\checkmark & \checkmark & \checkmark & \checkmark & 93.99 & 91.32 \\
		2 & & \checkmark & \checkmark & \checkmark & 93.36 & 90.42 \\
		3 & & & \checkmark & \checkmark & 91.87 & 87.85 \\
		\hline
		4 & & \checkmark & & \checkmark & 92.64 & 89.66 \\
		5 & & \checkmark & \checkmark & & 92.47 & 89.47 \\
		6 & & \checkmark & \checkmark$\dag$ & \checkmark$\dag$ & 93.03 & 90.06 \\
		\hline\hline
	\end{tabular}}%
	\caption{Parsing accuracy on PTB dev set for different variants of the full
	model.
	\texttt{INIT} refers to using pre-trained word embddings to initialize 
	${\bf E}^{\rm form}$.
	\texttt{POS} refers to using POS tags in token embeddings.
	\texttt{L2R} and \texttt{R2L} respectively indicate whether to use the left-to-right and
	right-to-left query components.
	$\dag$ means the query component drops soft headword embeddings when
constructing RNN hidden states.
\vspace{-3ex}
}
	\label{tab:exp_ablative}
\end{table}

Here we try to study the impact of using pre-trained word embeddings, POS tags, as well
as the bi-directional query components on our model.
First of all, we start from our best model (Model~1 in
Table\,\ref{tab:exp_ablative}) on English, which uses 300 as the
token embedding dimension and 368 as the hidden layer size.
We keep those model parameter dimensions unchanged and analyze different factors by comparing the parsing accuracy on PTB dev set.  

The results are summarized in Table\,\ref{tab:exp_ablative}.
Comparing Models~1--3, it can be observed that without using pre-trained word
embeddings, both UAS and LAS drop by 0.6\%, and without using POS tags in token
embeddings, the numbers further drop by 1.6\% in UAS and around 2.6\% in LAS.
In terms of query components, using single query component (Models~4--5)
degrades UAS by 0.7--0.9\% and LAS by around 1.0\%, compared with
Model~2.
For Model~6, the soft headword embedding is only used for arc label predictions
but not fed into the next hidden state, which is around 0.3\% worse than
Model~2.
This supports the hypothesis about the usefulness of the parsing history
information.
We also implement a variant of Model~6 which produces one $\avec_t$ instead two
by using both $\qvec_t^l$ and $\qvec_t^r$ in (\ref{eq:attention_score}).
It gets 92.44\% UAS and 89.26\% LAS, indicating that naively applying a
bi-directional RNN may not be enough.

\section{Related Work}

\paragraph{Neural Dependency Parsing:}
Recently developed neural dependency parsers are mostly transition-based models,
which read words sequentially from a buffer into a stack and incrementally build
a parse tree by predicting a sequence of transitions
\cite{Yamada2003IWPT,Nivre2003IWPT,Nivre2004ACL}.
A feed-forward neural network is used in \cite{Chen2014EMNLP}, where they
represent the current state with 18 selected elements such as the top words on the
stack and buffer.
Each element is encoded by concatenated embeddings of words, POS tags, and arc
labels.
Their dependency parser achieves improvement on both accuracy and parsing speed.
\newcite{Weiss2015ACL} improve the parser using semi-supervised structured
learning and unlabeled data.
The model is extended to integrate parsing and tagging in \cite{Alberti2015EMNLP}.
On the other hand, \newcite{Dyer2015ACL} develop the stack LSTM architecture,
which uses three LSTMs to respectively model the sequences of buffer states,
stack states, and actions.
Unlike the transition-based formulation, the proposed BiAtt-DP directly predicts
the headword and the dependency relation at each time step.
Specifically, there is no explicit representation of actions or headwords in our
model.
The model learns to retrieve the most relevant information from the input memory
to make decisions on headwords and head-modifier relations.

\paragraph{Graph-based Dependency Parsing:}
In addition to the transition-based parsers, another line of research in dependency
parsing uses graph-based models.
Graph-based parser usually build a dependency tree from a directed graph and
learns to scoring the possible arcs.
Due to this nature, non-projective parsing can be done straightforwardly by most
graph-based dependency parsers.
The MSTParser \cite{McDonald2005EMNLP} and the TurboParser
\cite{Martins2010EMNLP}
are two examples of graph-based parsers. 
The MSTParser formulates the parsing as searching for the MST, whereas the
TurboParser performs approximate variational inference over a factor graph.
The RBGParser proposed in \cite{Lei2014ACL} can also be viewed as a graph-based
parser, which scores arcs using low-dimensional continuous features derived
from low-rank tensors as well as features used by MSTParser/TurboParser.
It also employs a sampler-based algorithm for parsing \cite{ZhangLei2014ACL}.

\paragraph{Neural Attention Model:}
The proposed BiAtt-DP is closely related to the memory network
\cite{Sukhbaatar2015NIPS} for question answering, as well as the neural
attention models for machine translation \cite{Bahdanau2015ICLR} and
constituency parsing \cite{Vinyals2015NIPS}.
The way we query the memory component and obtain the soft headword embeddings is
essentially the attention mechanism.
However, different from the above studies where the alignment information is latent,
in dependency parsing, the arc between the modifier and headword is known during
training.
Thus, we can utilize these labels for attention weights.
The similar idea is employed by the pointer network in
\cite{VinyalsMeire2015NIPS}, which is used to solve three different
combinatorial optimization problems.

\section{Conclusion}
In this paper, we develop a bi-directional attention model by encouraging 
agreement between the latent attention alignments.
Through a simple and interpretable approximation, we make the connection between
latent and observed alignments for training the model.
We apply the bi-directional attention model incorporating the agreement
objective during training to the
proposed memory-network-based dependency parser.
The resulting parser is able to implicitly capture the high-order parsing
history without suffering from issue of high computational complexity for
graph-based dependency parsing.

We have carried out empirical studies over 14 languages.
The parsing accuracy of the proposed model is highly competitive with
state-of-the-art dependency parsers. 
For English, the proposed BiAtt-DP outperforms all graph-based parsers.
It also achieves state-of-the-art performance in 6 languages in terms of UAS,
demonstrating the effectiveness of the proposed mechanism of bi-directional
attention with agreement and its use in dependency parsing.

\appendix

\section{Upper Bound on $H^2(\pvec, \qvec)$}
\label{sec:appendix_ub}
Here, we use the following definition of squared Hellinger distance for
countable space
\begin{eqnarray*}
	H^2(\pvec, \qvec) & = & 
	{1 \over 2} \sum_i
	(\sqrt{\pvec_i} - \sqrt{\qvec_i})^2
\end{eqnarray*}
where $\pvec, \qvec \in \Delta^k$ are two $k$-simplexes.
Introducing $\gvec \in \Delta^k$, the squared Hellinger distance can be upper
bounded as
\begin{eqnarray}
	H^2(\pvec,\qvec)  &\leq&
	\label{eqn:1st_ub}
	\sqrt{2}H(\pvec, \qvec) \\
	\label{eqn:2nd_ub}
	&\leq& \sqrt{2} \left[
	H(\pvec, \gvec) + H(\qvec, \gvec)
	\right] \\
	\label{eqn:3nd_ub}
	&\leq& 2 \sqrt{ H^2(\pvec, \gvec) + H^2(\qvec, \gvec) }
\end{eqnarray}
where (\ref{eqn:1st_ub}), (\ref{eqn:2nd_ub}) and (\ref{eqn:3nd_ub}) follow
the inequalities between the $\ell_1$-norm and
the $\ell_2$-norm, the triangle inequality defined for a metric,
and the Cauchy-Schwarz's inequality, respectively.
Using the relationship between the KL-divergence and the squared Hellinger
distance, (\ref{eqn:3nd_ub}) can be further bounded by
\begin{eqnarray*}
	\label{eqn:4nd_ub} 
	2 \sqrt{ D(\gvec || \pvec) + D(\gvec || \qvec) }.
\end{eqnarray*}

\section{UAS Scores of MSTParsers}
\label{sec:appendix_uas_mst}
\begin{table}[h]
	\centering
	\begin{tabular}{l|cccc}
		\hline\hline
		{\bf Language}
		& \multicolumn{2}{c}{\bf 1st-order}
		& \multicolumn{2}{c}{\bf 2nd-order} \\
		\hline
		Arabic 			& 78.30 & (2.02) & 78.75 & (1.57) \\
		Bulgarian 	& 90.98 & (3.00) & 91.56 & (2.42) \\
		Czech 			& 86.18 & (4.88) & 87.30 & (3.76) \\
		Danish 			& 89.84 & (1.80) & 90.50 & (1.14) \\
		Dutch 			& 82.89 & (4.54) & 84.11 & (3.32) \\
		German 			& 89.54 & (3.17) & 90.14 & (2.57) \\
		Japanese 		& 93.38 & (0.14) & 92.92 & (0.60) \\
		Portuguese 	& 89.92 & (3.17) & 91.08 & (2.01) \\
		Slovene 		& 82.09 & (4.54) & 83.25 & (3.38) \\
		Spanish 		& 83.79 & (4.59) & 84.33 & (4.05) \\
		Swedish 		& 88.27 & (1.95) & 89.05 & (1.17) \\
		Turkish 		& 74.81 & (3.74) & 74.39 & (4.16) \\
		\hline
		{\bf Average} & 85.83 & (2.85) & 86.45 & (2.23) \\
		\hline\hline
	\end{tabular}
	\caption{UAS scores of 1st-order and 2-nd order MSTParsers on 12 languages in
		the CoNLL 2006 shared task \protect\cite{CONLL2006}.
		We use the numbers reported in \protect\cite{Lei2014ACL}.
		Numbers in brackets indicate the absolute improvement of the proposed
	BiAtt-DP over the MSTParsers.}
	\label{tab:exp_conll2006_mst}
\end{table}


\end{document}